\documentclass[11pt]{article}

\usepackage[preprint]{acl}

\usepackage{times}
\usepackage{lipsum}
\usepackage{comment}
\usepackage{latexsym}
\usepackage{amsmath}

\usepackage[T1]{fontenc}

\usepackage[utf8]{inputenc}

\usepackage{microtype}

\usepackage{inconsolata}

\usepackage{graphicx}

\usepackage{enumitem}
\usepackage{tcolorbox}
\usepackage{algorithm}
\usepackage[noend]{algpseudocode}
\title{Reducing Text Bias in Synthetically Generated MCQAs for VLMs in Autonomous Driving}

\author{
  Sutej Kulgod \quad
  Sean Ye \quad
  Sanchit Tanwar \quad
  Chris Heckman \\
  Zoox, Inc. \\
  \texttt{\{skulgod,sye,stanwar,checkman\}@zoox.com}
}

\begin{document}
\maketitle
\begin{abstract}
Multiple Choice Question Answering (MCQA) benchmarks are an established standard for measuring Vision Language Model (VLM) performance in driving tasks. However, we observe the known phenomenon that synthetically generated MCQAs are highly susceptible to \emph{hidden textual cues} that allow models to exploit linguistic patterns rather than visual context. Our results show that a VLM fine-tuned on such data can achieve accuracy comparable to human-validated benchmarks even without visual input. Our proposed method reduces blind accuracy from +66.9\% above random to +2.9\%, eliminating the vast majority of exploitable textual shortcuts. By decoupling the correct answer from linguistic artifacts and employing a curriculum learning strategy, we force the model to rely on visual grounding, ensuring that performance accurately reflects perceptual understanding. 
\end{abstract}

\section{Introduction}

Vision Language Models (VLMs) have become integral to the modern autonomous driving stack \citep{nvidia2025alpamayor1,hwang2025emmaendtoendmultimodalmodel}, especially for tackling safety-critical ``long-tail'' scenarios. While benchmarks fall into free-form QA \citep{park2025nuplanqa, sima2023drivelm, malla2023drama} or Multiple Choice Question Answering (MCQA) \citep{xie2025drivebench, khalili2025autodriveqa,tong2025dvbench,lu2024idkb}, MCQA is often preferred due to its straightforward scoring and structured output.

However, the integrity of these benchmarks depends on the quality of distractors. Recent works have explored automated MCQA creation \citep{zhang2025challengingmcq,alhazmi2024distractor,loginova2025addressing}, but these methods do not guarantee the absence of \emph{hidden textual cues}. Our investigation reveals that when VLMs generate both the correct answer and distractors for a driving task, they introduce subtle linguistic markers. We find that small VLMs trained on this data quickly learn to ignore visual inputs, achieving high accuracy by merely processing text. This is a form of shortcut learning that invalidates the benchmark as a measure of the trained VLM's generalizability.

The core contributions to this work are:
\begin{itemize}[nosep]
\item A framework for identifying hidden textual bias using trained VLMs with \emph{zeroed-out visual inputs} during inference.
\item A \emph{two-stage MCQA generation} method, that first creates the question-answer pairs for each sample and second, samples distractors from answers of other samples, reducing linguistic bias towards the correct option.
\item A \emph{curriculum learning approach} that drops MCQ options early in training to encourage stronger visual-textual grounding.
\end{itemize}

\begin{figure*}[ht]
    \centering
  \includegraphics[width=0.95\textwidth]{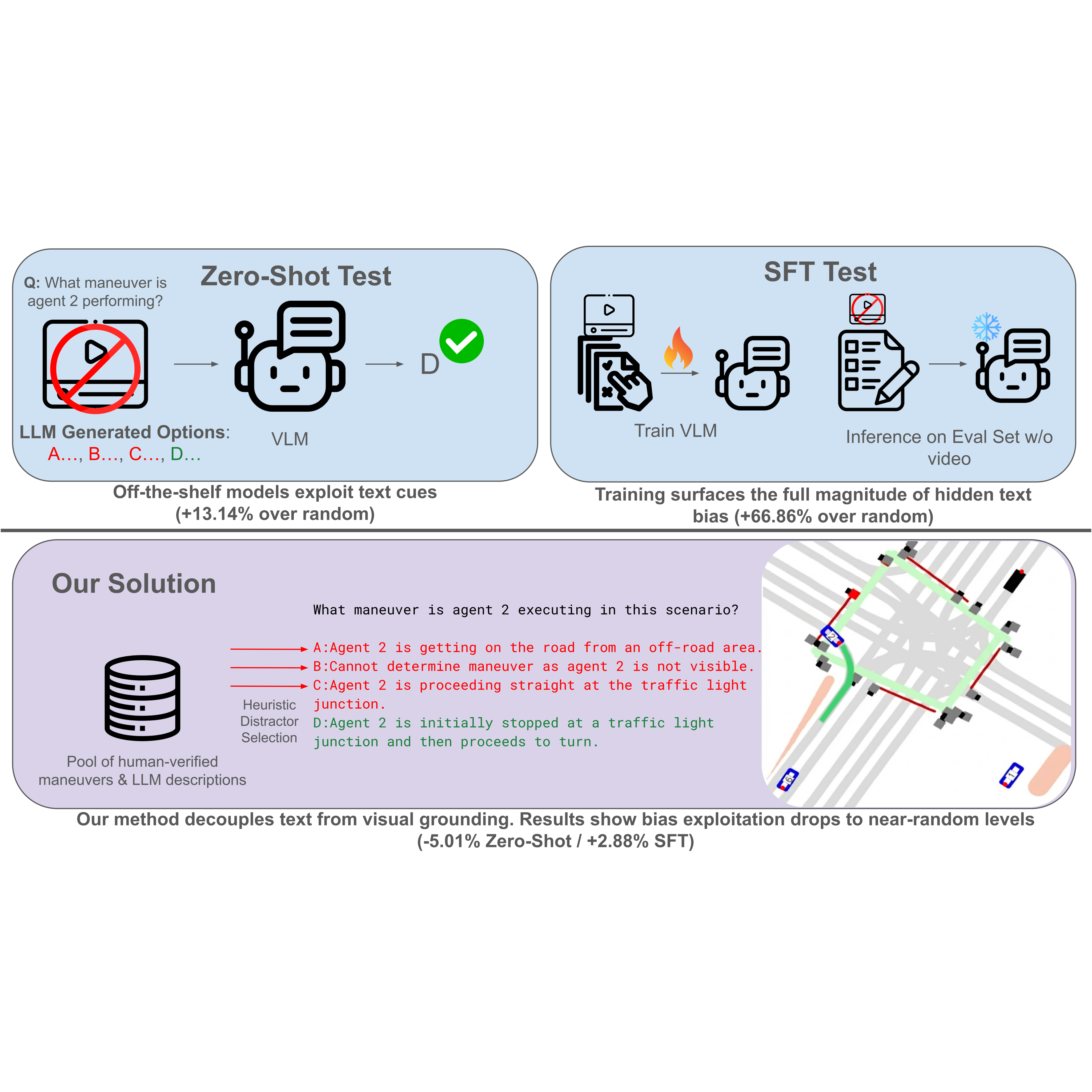}
  \caption {\textbf{Top}: We employ two video-disabled evaluations to detect textual cues. The Zero-Shot Test reveals inherent linguistic patterns in off-the-shelf models, while the SFT Test exposes the full magnitude of shortcut learning when a model is trained on biased data. \textbf{Bottom}: Our method replaces LLM-generated distractors with real descriptions sampled from elsewhere in the dataset. This reduces bias exploitation in both Zero-Shot and SFT.}
  \label{fig:overview}
  \vspace{-5mm}
\end{figure*}

\section{Related Work}


Foundation models for driving \cite{hwang2025emmaendtoendmultimodalmodel} have gained popularity because they leverage internet-scale data to develop ``physical common sense" and emergent reasoning abilities, allowing them to perform scene understanding by answering natural-language questions to handle the safety-critical ``long-tail" scenarios that traditional modular systems struggle with. This usage encourages robust benchmarks to establish industry standards, validate safety in rare corner cases, and provide an objective measure of progress. While early benchmarks used graph-based VQA like DriveLM \cite{sima2023drivelm}, modern frameworks like NuPlanQA \cite{park2025nuplanqa}, DriveBench \cite{xie2025drivebench}, DriveLM \cite{sima2023drivelm}, and AutoDrive-QA \cite{khalili2025autodriveqa} provide QA datasets at scale for supervised fine-tuning (SFT) and testing. These datasets however often lack an analysis of how synthetic MCQ options might introduce textual cues, allowing models to succeed without grounding their reasoning in the vision inputs. 

The integrity of multiple-choice evaluation depends on the quality and plausibility of distractors. Modern agentic frameworks like AutoConverter \cite{zhang2025AutoConverter}, AutoDrive-QA \cite{khalili2025autodriveqa}, GOBBET \cite{lu2022good} use language models to generate hard distractors. Many previous works have looked specifically into MCQA creation for LLM evaluation \citep{zhang2025challengingmcq,alhazmi2024distractor,loginova2025addressing}. While most of these focus on text only data, a recent work has shed light into hidden biases for VLMs \cite{loginova2025addressing}. In our work we find that there are more hidden textual cues that get exposed when we train a small model using this MCQ data. To address this, we propose a distractor selection workaround designed to decouple task-related signals from these textual artifacts.

\section{Problem Setup and Synthetic MCQA Generation}
In this section, we formalize synthetic MCQA construction for autonomous driving, defining the base driving dataset, the target MCQA format, and the data generation pipeline.

\subsection{Driving Dataset Creation}
\label{sec:task_def}
We assume access to a labeled driving dataset $\mathcal{D}_{base} = \{(v_i, y_i)\}_{i=1}^N$, where $v_i$ denotes a 10s birds-eye-view (BEV) video clip depicting a driving scene and $y_i \in \mathcal{Y}$ is the maneuver label describing the behavior of a selected non-ego agent in the scene. Maneuver labels are uniformly balanced and correspond to semantically meaningful agent behaviors such as lane changes, turns, etc. as defined in Appendix \ref{sec:appendix} or \emph{agent not visible} when the selected agent is not in $v_i$.





From the base dataset $\mathcal{D}_{base}$, we construct a multiple-choice question answering (MCQA) task where each sample requires identifying the maneuver performed by a target agent in a driving scene. An MCQA instance is defined as $m_i = (q_i, v_i, \mathcal{A}_i)$, where $q_i$ is a natural-language question referring to the target agent, $v_i$ is the associated BEV video clip, and $\mathcal{A}_i$ is a set of answer options. Each answer set $\mathcal{A}_i$ contains one correct ground-truth answer $a_i^{gt}$ and a set of plausible but visually incorrect distractors $\mathcal{A}_i^{cand} = \mathcal{A}_i \setminus {a_i^{gt}}$. Throughout our experiments we use a fixed four-way multiple-choice format.


\paragraph{Stage I: Natural Language Formatting of Ground-Truth Answer}

The first stage of synthetic MCQA construction produces a natural-language question–answer pair from maneuver label $y_i$ for each sample. Given a base dataset entry $(v_i, y_i)$, the question–answer pair is generated as $(q_i, a_i^{gt}) = f_{NL}(v_i, y_i)$, where $f_{NL}$ is an expert model that maps structured maneuver labels to textual descriptions. The expert model (Gemini~2.5 for our experiments) is restricted from introducing semantic content beyond the ground-truth label. 



\paragraph{Stage II (Baseline): LLM-Based Distractor Generation}
In the second stage of synthetic MCQA construction, we populate the answer set $\mathcal{A}_i$ by generating candidate distractors for each question-video-answer triple $(q_i, v_i, a_i^{gt})$ produced in Stage~I. Distractors are generated using the same expert model employed during answer realization.

Formally, the answer set is constructed as $\mathcal{A}_i = \{a_i^{gt}\} \cup \mathcal{A}_i^{cand}$, where the candidate distractors are sampled as $\mathcal{A}_i^{cand} \sim p_{\mathrm{LLM}}(\cdot \mid q_i, v_i, a_i^{gt})$. The resulting distractors are intended to be semantically plausible maneuver descriptions that do not match the visual content of $v_i$. Applying this two-stage procedure across the base dataset yields a synthetic MCQA dataset $\mathcal{D}_{\mathrm{llm}} = \{(q_i, v_i, \mathcal{A}_i)\}_{i=1}^N$.

\begin{table}[t]
  \centering
  \begin{tabular}{llp{2.5cm}p{1.5cm}p{1.5cm}}
    \hline
    \textbf{Dataset} & \textbf{Model} & \textbf{Accuracy \%}\\
    \hline
    $D_{llm}$ & Gemini 2.5 Pro & 38.14  (+13.14) \\
    $D_{llm}$ & Gemini 2.5 flash & 32.73  (+7.73)\\
    $D_{llm}$ & Qwen2-VL-7B & 29.56 (+4.56)\\
    \hline
    $D_{new}$ & Gemini 2.5 Pro & 19.99 (-5.01)\\
    $D_{new}$ & Gemini 2.5 flash & 19.88 (-5.12)\\
    $D_{new}$ & Qwen2-VL-7B & 21.38 (-4.62)\\
    \hline
  \end{tabular}
  \caption{\label{tab:sanitycheck}
    Results obtained by passing the MCQ through LLMs without video context. We present results as \emph{Overall accuracy (Accuracy above random)}
  }
  \vspace{-5mm}
\end{table}
\subsection{Empirical Analysis of Synthetic MCQs}
\label{sec:empirical_analysis}
We perform a series of analyses to assess whether the synthetically generated dataset $D_{\mathrm{llm}}$ exhibits unintended textual biases that allow models to perform well without visual grounding. Our analysis evaluates both off-the-shelf VLMs and models trained directly on the synthetic MCQs.
\paragraph{Zero-Shot Test}
We conduct a set of diagnostic tests designed to detect superficial cues in the question or answer options. We verify that the position of the correct answer within the answer set is uniformly distributed and manually inspect a random subset of MCQs to ensure there are no obvious textual giveaways (such as distractor lengths or jargon). We then evaluate multiple pretrained VLMs on $D_{\mathrm{llm}}$ using only textual inputs $(q_i, \mathcal{A}_i)$ while withholding the video $v_i$. As shown in Table~\ref{tab:sanitycheck}, several models achieve accuracy above random chance in this setting, with the model used for data generation exhibiting the strongest performance. This suggests the presence of model-specific linguistic regularities embedded in the synthetic distractors.




\paragraph{Supervised Fine-tuned Test}
\begin{table*}[t]
\small
  \centering
  \begin{tabular}{llp{1.4cm}p{1.4cm}p{1.4cm}p{1.4cm}p{2.5cm}}
    \hline
    \textbf{Dataset} & \textbf{Training Strategy} & \multicolumn{2}{c}{\textbf{Accuracy (\%) $\uparrow$}} & \multicolumn{3}{c}{\textbf{Accuracy Video Zeroed Out (\%)}}\\
    & & Normal & Shuffled & ($D_{NV}$) $\uparrow$ & ($D_{N}$) $\downarrow$ & ($D_{V}$) $\downarrow$\\
    \hline
    $D_{llm}$ & No Training & 29.23\% & 6.46\% & 17.48\% & 32.22\% & 23.4\% (-1.6\%) \\
    $D_{llm}$ & Regular SFT & 93.82\% & 89.03\% & 99.68\% & 64.66\% & 91.9\% (+66.9\%)\\
    \hline
    $D_{new}$ & No Training & 24.38\% & 6.23\% & 12.94\% & 29.85\% & 23.1\% (-1.9\%)\\
    $D_{new}$ & Regular SFT & 75.72\% & 66.43\% & \textbf{100\%} & 0.21\% & \textbf{27.9\% (+2.9\%)} \\
    $D_{new}$ & Curriculum & 77.28\% & 70.82\% & \textbf{100\%} & 0.84\% & 30.4\% (+5.4\%)\\
    $D_{new}$ & Regular SFT (FF) & 88.08\% & 84.35\% & \textbf{100\%} & 0.21\% & 29.5\% (+4.5\%)\\
    $D_{new}$ & Curriculum (FF) & \textbf{92.09\%} & \textbf{87.64\%} & \textbf{100\%} & \textbf{0.0\%} & 29.9\% (+4.9\%)\\
    \hline
  \end{tabular}
  \caption{\label{tab:results}
    Qwen2-VL-2B training and evaluation results on different datasets and training strategies. Models are trained with ViT and projector frozen unless noted as full fine-tune \emph{(FF)}. Model evaluation uses greedy sampling ($k=1$) to eliminate variance. For $D_V$ we present results as \emph{Overall accuracy (Accuracy above random)}.
  }
\end{table*}
To uncover biases that emerge during training, we fine-tune a Qwen2-VL-2B model~\cite{Qwen2-VL} on $D_{\mathrm{llm}}$ and evaluate it with shuffled answer ordering~\cite{tong2025dvbench} and zeroed-out visual inputs. For the zeroed-video setting, we partition the test set into $D_{NV}$ (agent not visible is correct), $D_{N}$ (agent not visible is an incorrect option), and $D_{V}$ (agent not visible option absent). This split isolates distinct failure modes: $D_{NV}$ tests whether the model can identify agent absence, while $D_{N}$ and $D_{V}$ probe reliance on textual cues when no visual evidence is available. Despite high nominal accuracy after fine-tuning (93.82\%), performance remains high without video input; in particular, accuracy on $D_{V}$ and $D_{N}$ reaches 91.86\% and 64.66\% respectively (Table~\ref{tab:results}), indicating that the model recovers the correct answer based on the options' linguistic structure rather than a semantic mismatch between the distractor and the multimodal inputs.

\section{Methodology}
\label{subsec:overcomehiddenbias}
To eliminate text bias introduced during distractor construction, we modify the Stage~II of synthetic MCQA generation by decoupling distractor selection from VLMs. In the baseline dataset $D_{\mathrm{llm}}$, candidate answers $\mathcal{A}_i^{\mathrm{cand}}$ are generated by an LLM conditioned on the ground-truth answer $a_i^{\mathrm{gt}}$. In contrast, our proposed approach samples distractors in maneuver label space and reuses ground-truth answers from other samples.

After Stage~I, each MCQA instance $m_i$ is associated with a ground-truth maneuver label $y_i \in \mathcal{Y}$. For a given MCQA instance $m_i$, we construct the candidate answer set by sampling $K-1$ distractor labels $\{y_j\}$ such that $y_j \in \mathcal{Y} \setminus \{y_i\}$. For each distractor label $y_j$ we sample a ground-truth answer ${a_k}^{\mathrm{gt}}$ such that $y_k = y_j$. To ensure contextual consistency, agent identifiers in these answers are rewritten to match the target agent referenced in the question. Applying this heuristic across the dataset yields a debiased MCQA dataset $D_{\mathrm{new}}$, in which ground-truth answers and distractors are no longer coupled through a shared generative model.


\paragraph{Curriculum-Based Option Dropping}
In addition to modifying distractor construction, we employ a curriculum-based training strategy to further discourage reliance on textual cues. During supervised fine-tuning, answer options are randomly dropped from a fraction of training samples, converting MCQA instances $(q_i, v_i, \mathcal{A}_i)$ into open-ended question $(q_i, v_i)$ with target output answer $a_i^{\mathrm{gt}}$. This forces the model to generate the correct answer conditioned on visual input rather than selecting from a fixed option set $\mathcal{A}_i$.

We schedule the fraction of option dropping $x(t)$ for training step $t$ according to
\begin{equation}
\label{eq:curriculum}
x(t) = \max\!\left(d_{\min},\, d_{\max} - d_{\min}\left(t/\tau\right)^2\right),
\end{equation}
where $\tau$ is the number of training steps over which the curriculum is applied, and $d_{\min}$ and $d_{\max}$ denote the minimum and maximum drop ratios, respectively. Early in training, options are dropped aggressively to encourage visual grounding, and are gradually reintroduced as training progresses.

\section{Results \& Discussion}
We evaluate whether the debiased dataset $D_{\mathrm{new}}$ eliminates textual shortcuts under zero-shot and supervised fine-tuning (SFT) settings. Unlike the baseline $D_{\mathrm{llm}}$, models evaluated on $D_{\mathrm{new}}$ perform near chance without visual input.

\paragraph{Zero-shot evaluation}
We first assess textual bias using pretrained VLMs evaluated without video input. Models evaluated on $D_{\mathrm{new}}$ achieve accuracy close to random guessing (-5.01\%), in contrast to $D_{\mathrm{llm}}$ (+13.14\%), where several models perform significantly above random chance (Table~\ref{tab:sanitycheck}). Notably, the model used to generate the baseline dataset exhibits the strongest bias on $D_{\mathrm{llm}}$, while its performance drops below random on $D_{\mathrm{new}}$. This confirms that heuristic distractor sampling in maneuver label space removes model-specific linguistic cues that are exploitable in zero-shot settings.

\paragraph{Supervised fine-tuning evaluation}
We next evaluate whether textual bias re-emerges after training by fine-tuning a Qwen2-VL-2B model under controlled conditions. When trained on $D_{\mathrm{llm}}$, the model achieves high accuracy (93.82\%) and retains near-human performance even when the video is zeroed out (+66.86\%). In contrast, models trained on $D_{\mathrm{new}}$ exhibit substantially reduced performance without visual input (Table~\ref{tab:results}), particularly on the $D_V$ subset, where accuracy remains close to random (+2.88\% for regular SFT). The performance on $D_{N}$ subset being close to 0\% (+0.21\% for regular SFT) and $D_{NV}$ subset being 100\% also indicate that the model can no longer recover the correct answer from textual patterns alone and must rely on visual evidence to succeed.

\paragraph{Effect of curriculum-based training}
We further observe that curriculum-based training improves robustness. Compared to regular SFT, curriculum training yields higher accuracy while reducing sensitivity to answer shuffling and video removal. This suggests improved visual-textual grounding rather than increased reliance on option structure. The strongest performance is achieved when curriculum learning is combined with full fine-tuning of the vision encoder and projector, reflecting the importance of adapting visual representations for BEV inputs, which are uncommon in standard VLM pretraining. 
High performance on $D_{\mathrm{new}}$ therefore reflects genuine visual understanding rather than linguistic bias, validating the effectiveness of the proposed debiasing approach.

\section{Conclusion}
In this work, we systematically demonstrate that synthetically generated MCQAs are subject to hidden textual cues introduced by the VLMs used to generate them. We propose a distractor generation method that reduces this bias and a curriculum learning modification to improve grounding when using such data in training mixes. Our results confirm that these methods force models to rely on visual information rather than linguistic shortcuts.

\section*{Limitations}

We limit our analysis in this paper to training the Qwen2-VL-2B model for the maneuver classification task due to resource constraints. Training on more datasets and models can provide a deeper understanding about the hidden patterns and the conditions where they are more evident. Investigating similarity scoring like BLEU, word frequencies or embedding space distance for MCQAs can also uncover biases in the Zero-Shot test before subjecting models to VLM training. In our work we keep the zero-shot tests simple while leaving the majority of bias detection to model training as our main goal was to train a model to answer driving related questions in addition to developing a reliable benchmark. 

In our baseline experiments we use BEV videos with heuristic maneuver labels to generate MCQAs. The expert models might have limited exposure to such data resulting in the text patterns. Reproducing the experiments with a more common video classification tasks that the expert model has exposure to can give a more complete insight.



\section*{Acknowledgments}
We thank Nick Dulchin, Alejandro Galindo, Nicolas Hammer, Aaron Huang, Haresh Karnan, Mayank Ketkar, Siddarth Kodwani, Ashvini Pavaskar, Doron Portnoy, Jim Robinson-Bohnslav, Anindya Saha, Peter Schleede, Vaibhav Sinha, Siva Subramanian, Paul Viola, Jackson Waschura for their constructive conversations and valuable feedback. Generative AI tools were used for writing code for the experiments and visualizations. All sections were written with proofreading and grammatical reformatting from Gemini and ChatGPT. Gemini also supported the related work search specifically for MCQA generation and distractor quality evaluation in Section 2.

\bibliography{custom}

@INPROCEEDINGS{zhang2025challengingmcq,
  author={Zhang, Yuhui and Su, Yuchang and Liu, Yiming and Wang, Xiaohan and Burgess, James and Sui, Elaine and Wang, Chenyu and Aklilu, Josiah and Lozano, Alejandro and Wei, Anjiang and Schmidt, Ludwig and Yeung-Levy, Serena},
  booktitle={2025 IEEE/CVF Conference on Computer Vision and Pattern Recognition (CVPR)}, 
  title={Automated Generation of Challenging Multiple-Choice Questions for Vision Language Model Evaluation}, 
  year={2025},
  volume={},
  number={},
  pages={29580-29590},
  doi={10.1109/CVPR52734.2025.02754},
  summary={
  \begin{itemize}
    \item  The paper introduces AutoConverter which converts open ended VQAs to MCQs using GPT-4o
    \item They introduce VMC Bench using 20 VQA datasets into a unified MCQ dataset with ~9k questions.  This I think can be used as one of the datasets we test our hypothesis on?
    \item They use a convoluted web of “agents” to go from VQA to MCQs.
    \item The approach tests accuracy of pretrained models on test sets created by this method zero-shot. We have a different issue where we use the same approach to create the train and test set. 
    \item No major digging into the answer structure, word count or anything else. They do answer shuffling though to ensure that the metrics do not change. 
  \end{itemize}


    Interesting snippets:
    “we randomly sampled up to 500 questions, as recent studies suggest this sample size may suffice for evaluating model performance [44, 45],”
    “We choose N = 3 in our work because 4-choice is the most common configuration for multiple-choice questions and minimizes the risk of option selection bias for language models [63].”
    }
  }

@inproceedings{alhazmi2024distractor,
    title = "Distractor Generation in Multiple-Choice Tasks: A Survey of Methods, Datasets, and Evaluation",
    author = "Alhazmi, Elaf  and
      Sheng, Quan Z.  and
      Zhang, Wei Emma  and
      Zaib, Munazza  and
      Alhazmi, Ahoud",
    editor = "Al-Onaizan, Yaser  and
      Bansal, Mohit  and
      Chen, Yun-Nung",
    booktitle = "Proceedings of the 2024 Conference on Empirical Methods in Natural Language Processing",
    month = nov,
    year = "2024",
    address = "Miami, Florida, USA",
    publisher = "Association for Computational Linguistics",
    url = "https://aclanthology.org/2024.emnlp-main.799/",
    doi = "10.18653/v1/2024.emnlp-main.799",
    pages = "14437--14458",
    summary = "
    * The paper is a survey on distraction generation approaches, this can be a great resource to help write our relevant papers and intro sections. 
    * They have a small set of multi-modal datasets that are considered but they span different domains. 
    * The paper talks about distraction generation done using language model prompting using different approaches. However it does not go into the details of the issues seen and why some papers pick the method they do.
    * Good resource for distraction generation evaluation methods. 
    * N-gram metrics for evaluating overlap between GT distractors and generated distractors. It does not seem to be between correct and wrong answers. 
    * “like hallucination issues in PLMs (Ji et al., 2023) and a heavy reliance on costly human-annotated labels (Qu et al., 2024). To control this task generation (Zhang et al., 2023a), reinforcement learning from human feedback (RLHF) (Ouyang et al., 2022) and few-shot examples (Bitew et al., 2023) may be utilized to improve the trustworthiness of DG.”
"
}

@inproceedings{loginova2025addressing,
    title = "Addressing Blind Guessing: Calibration of Selection Bias in Multiple-Choice Question Answering by Video Language Models",
    author = "Loginova, Olga  and
      Bezrukov, Oleksandr  and
      Shekhar, Ravi  and
      Kravets, Alexey",
    editor = "Che, Wanxiang  and
      Nabende, Joyce  and
      Shutova, Ekaterina  and
      Pilehvar, Mohammad Taher",
    booktitle = "Proceedings of the 63rd Annual Meeting of the Association for Computational Linguistics (Volume 1: Long Papers)",
    month = jul,
    year = "2025",
    address = "Vienna, Austria",
    publisher = "Association for Computational Linguistics",
    url = "https://aclanthology.org/2025.acl-long.162/",
    doi = "10.18653/v1/2025.acl-long.162",
    pages = "3216--3246",
    ISBN = "979-8-89176-251-0"
}

@inproceedings{xie2025drivebench,
        author    = {Xie, Shaoyuan and Kong, Lingdong and Dong, Yuhao and Sima, Chonghao and Zhang, Wenwei and Chen, Qi Alfred and Liu, Ziwei and Pan, Liang},
        title     = {Are VLMs Ready for Autonomous Driving? An Empirical Study from the Reliability, Data and Metric Perspectives},
        booktitle = {Proceedings of the IEEE/CVF International Conference on Computer Vision (ICCV)},
        month     = {October},
        year      = {2025},
        pages     = {6585-6597}
}

@inproceedings{tong2025dvbench,
author = {Zeng, Tong and Wu, Longfeng and Shi, Liang and Zhou, Dawei and Guo, Feng},
title = {Are Vision LLMs Road-Ready? A Comprehensive Benchmark for Safety-Critical Driving Video Understanding},
year = {2025},
isbn = {9798400714542},
publisher = {Association for Computing Machinery},
address = {New York, NY, USA},
url = {https://doi.org/10.1145/3711896.3737396},
doi = {10.1145/3711896.3737396},
booktitle = {Proceedings of the 31st ACM SIGKDD Conference on Knowledge Discovery and Data Mining V.2},
pages = {5972–5983},
numpages = {12},
location = {Toronto ON, Canada},
series = {KDD '25}
}

@article{park2025nuplanqa,
  title={Nuplanqa: A large-scale dataset and benchmark for multi-view driving scene understanding in multi-modal large language models},
  author={Park, Sung-Yeon and Cui, Can and Ma, Yunsheng and Moradipari, Ahmadreza and Gupta, Rohit and Han, Kyungtae and Wang, Ziran},
  journal={arXiv preprint arXiv:2503.12772},
  year={2025}
}

@misc{khalili2025autodriveqa,
      title={AutoDrive-QA: A Multiple-Choice Benchmark for Vision-Language Evaluation in Urban Autonomous Driving}, 
      author={Boshra Khalili and Andrew W. Smyth},
      year={2025},
      eprint={2503.15778},
      archivePrefix={arXiv},
      primaryClass={cs.CV},
      url={https://arxiv.org/abs/2503.15778}, 
}

@misc{nvidia2025alpamayor1,
      title={Alpamayo-R1: Bridging Reasoning and Action Prediction for Generalizable Autonomous Driving in the Long Tail}, 
      author={NVIDIA and : and Yan Wang and Wenjie Luo and Junjie Bai and Yulong Cao and Tong Che and Ke Chen and Yuxiao Chen and Jenna Diamond and Yifan Ding and Wenhao Ding and Liang Feng and Greg Heinrich and Jack Huang and Peter Karkus and Boyi Li and Pinyi Li and Tsung-Yi Lin and Dongran Liu and Ming-Yu Liu and Langechuan Liu and Zhijian Liu and Jason Lu and Yunxiang Mao and Pavlo Molchanov and Lindsey Pavao and Zhenghao Peng and Mike Ranzinger and Ed Schmerling and Shida Shen and Yunfei Shi and Sarah Tariq and Ran Tian and Tilman Wekel and Xinshuo Weng and Tianjun Xiao and Eric Yang and Xiaodong Yang and Yurong You and Xiaohui Zeng and Wenyuan Zhang and Boris Ivanovic and Marco Pavone},
      year={2025},
      eprint={2511.00088},
      archivePrefix={arXiv},
      primaryClass={cs.RO},
      url={https://arxiv.org/abs/2511.00088}, 
}

@inproceedings{zhang2025AutoConverter,
  title={Automated Generation of Challenging Multiple-Choice Questions for Vision Language Model Evaluation},
  author={Yuhui Zhang and Yuchang Su and Yiming Liu and Xiaohan Wang and James Burgess and Elaine Sui and Chenyu Wang and Josiah Aklilu and Alejandro Lozano and Anjiang Wei and Ludwig Schmidt and Serena Yeung-Levy},
  booktitle={Conference on Computer Vision and Pattern Recognition (CVPR)},
  year={2025}
}

@inproceedings{lu2022good,
  title={Good, better, best: Textual distractors generation for multiple-choice visual question answering via reinforcement learning},
  author={Lu, Jiaying and Ye, Xin and Ren, Yi and Yang, Yezhou},
  booktitle={Proceedings of the IEEE/CVF Conference on Computer Vision and Pattern Recognition},
  pages={4921--4930},
  year={2022}
}

@misc{hwang2025emmaendtoendmultimodalmodel,
      title={EMMA: End-to-End Multimodal Model for Autonomous Driving}, 
      author={Jyh-Jing Hwang and Runsheng Xu and Hubert Lin and Wei-Chih Hung and Jingwei Ji and Kristy Choi and Di Huang and Tong He and Paul Covington and Benjamin Sapp and Yin Zhou and James Guo and Dragomir Anguelov and Mingxing Tan},
      year={2025},
      eprint={2410.23262},
      archivePrefix={arXiv},
      primaryClass={cs.CV},
      url={https://arxiv.org/abs/2410.23262}, 
}

@article{sima2023drivelm,
    title={DriveLM: Driving with Graph Visual Question Answering},
    author={Sima, Chonghao and Renz, Katrin and Chitta, Kashyap and Chen, Li and Zhang, Hanxue and Xie, Chengen and Luo, Ping and Geiger, Andreas and Li, Hongyang},
    journal={arXiv preprint arXiv:2312.14150},
    year={2023}
}

@misc{lu2024idkb,
      title={Can LVLMs Obtain a Driver's License? A Benchmark Towards Reliable AGI for Autonomous Driving}, 
      author={Yuhang Lu and Yichen Yao and Jiadong Tu and Jiangnan Shao and Yuexin Ma and Xinge Zhu},
      year={2024},
      eprint={2409.02914},
      archivePrefix={arXiv},
      primaryClass={cs.CV},
      url={https://arxiv.org/abs/2409.02914}, 
}

@inproceedings{malla2023drama,
  author={Malla, Srikanth and Choi, Chiho and Dwivedi, Isht and Hee Choi, Joon and Li, Jiachen},
  booktitle={2023 IEEE/CVF Winter Conference on Applications of Computer Vision (WACV)}, 
  title={DRAMA: Joint Risk Localization and Captioning in Driving}, 
  year={2023},
  volume={},
  number={},
  pages={1043-1052},
  doi={10.1109/WACV56688.2023.00110}}

@article{Qwen2-VL,
  title={Qwen2-VL: Enhancing Vision-Language Model's Perception of the World at Any Resolution}, 
  author={Peng Wang and Shuai Bai and Sinan Tan and Shijie Wang and Zhihao Fan and Jinze Bai and Keqin Chen and Xuejing Liu and Jialin Wang and Wenbin Ge and Yang Fan and Kai Dang and Mengfei Du and Xuancheng Ren and Rui Men and Dayiheng Liu and Chang Zhou and Jingren Zhou and Junyang Lin},
  journal={arXiv preprint arXiv:2409.12191},
  year={2024}
}

\appendix

\section{Appendix}
\label{sec:appendix}
\subsection{Maneuver Types}
The maneuver determination for selected agents is done  using heuristic conditions on their longitudinal and lateral movement on the road in a given scene. The maneuver types used for data generation and grouping for distractor section are as follows: 
\begin{enumerate}[nosep]
    \item \textbf{STRAIGHT} - The agent is going straight
    \item \textbf{TURNING} - The agent is making a turn in the junction
    \item \textbf{NUDGE AROUND OBSTACLE} - The agent is going around an obstacle like Vehicle, pedestrian, garbage etc..
    \item \textbf{LANE CHANGE} - The agent is making a lane change
    \item \textbf{REVERSING} - The agent is moving backwards on the road
    \item \textbf{U TURN} - The agent is making a U turn
    \item \textbf{GETTING ON ROAD} - The agent is transitioning from being off road to being on road
    \item \textbf{PARKING LANE CUTIN} - The agent is transitioning from being in the parking lane to being in the driving lane
    \item \textbf{STATIONARY} - The agent is parked or double parked
    \item \textbf{GOING OFF ROAD} - The agent is transitioning from being on road to being off road
    \item \textbf{STOPPED} - The agent is stopped for traffic light or stop sign
\end{enumerate}

\subsection{Dataset Details}
\begin{figure}[t]
    \centering
    \includegraphics[width=1\linewidth]{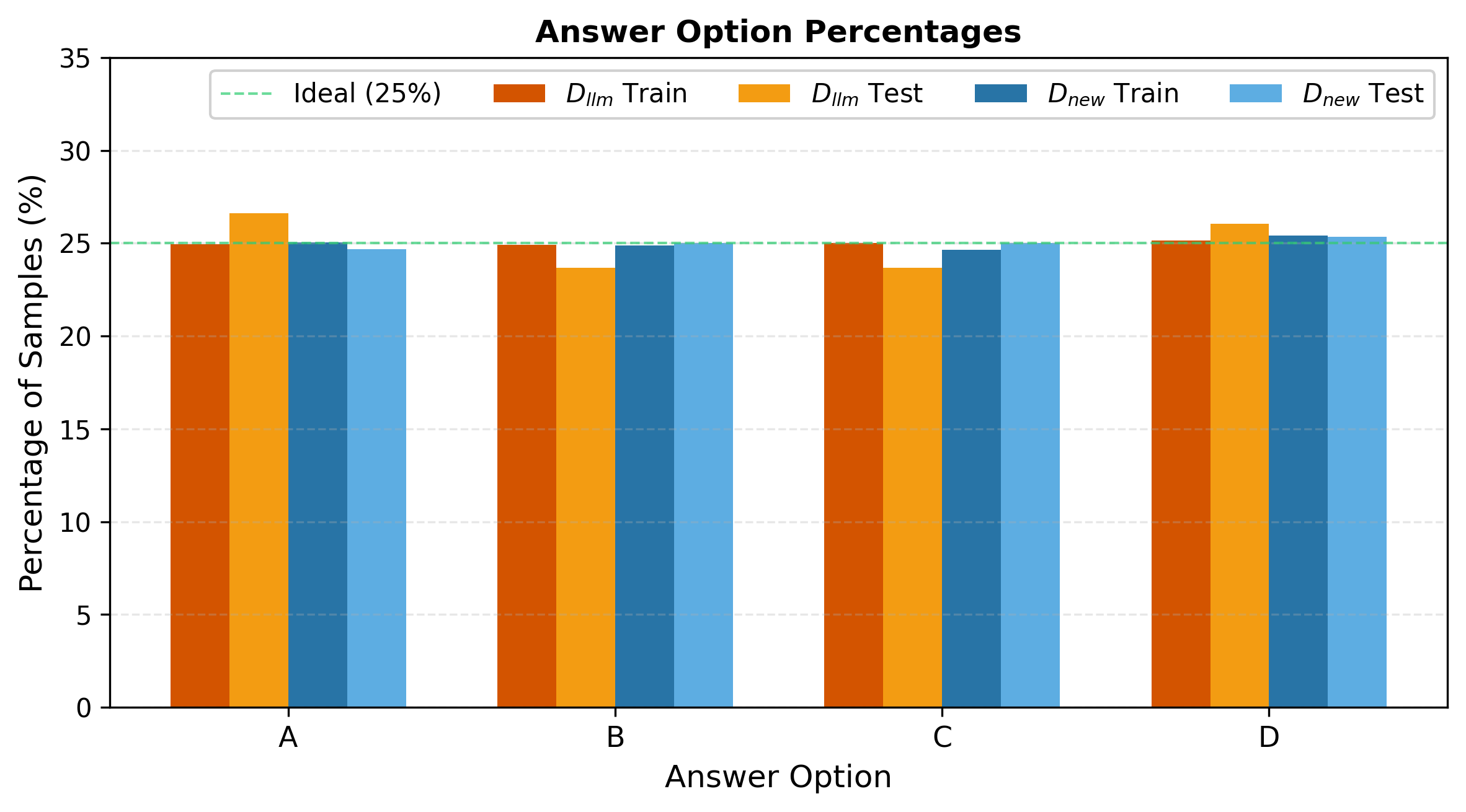}
    \caption{Distribution of the correct option in the train and test subsets of $D_{llm}$ and $D_{new}$. We see even distribution of correct answers for all datasets.}
    \label{fig:answer_distribution}
\end{figure}

We use the same samples $D_{base}$ and Stage~I processing for $D_{llm}$ and $D_{new}$. The driving dataset $D_{base}$ contains 59940 samples, out of which, we hold out 1796 samples for the test set. We hold out the same samples for  $D_{llm}$ and $D_{new}$ to keep the accuracy numbers comparable. We begin with base dataset $D_{base}$ containing close to uniform label distribution for the 12 maneuver types resulting in ~7395 (8.33\%) samples per maneuver type and we maintain the uniformity for the holdout set as well by sampling 150 or 149 samples per maneuver type. We run an automated pipeline with the BEV videos to detect when the agent in question is not visible in it and replace the base maneuver label with \emph{agent not visible} label. This results in forming the final $D_{base}$ that is used for data generation, with 18.8\% of the data being marked as \emph{agent not visible} while keeping rest of the maneuver labels close to uniform, ranging between 6\% to 8\% of the data distribution as shown in Table \ref{tab:maneuver_distribution}. We also ensure that the label distribution is close to uniform in our train and test datasets to avoid model bias as seen in Figure \ref{fig:answer_distribution}.

\begin{table}[t]
  \centering
  \begin{tabular}{lp{1.9cm}p{1.9cm}}
    \hline
    \textbf{Maneuver Type} & \textbf{Original (\%)} & \textbf{Processed (\%)} \\
    \hline
    Not Visible & --- & 18.82 \\
    \hline
    Reversing & 8.34 & 7.89 \\
    Turning & 8.34 & 7.13 \\
    Nudge Around & 8.32 & 7.13 \\
    Hard Stopped & 8.33 & 6.88 \\
    Going Off Road & 8.34 & 6.82 \\
    Getting On Road & 8.32 & 6.68 \\
    U-Turn & 8.34 & 6.61 \\
    Straight & 8.34 & 6.60 \\
    Parking Cut-in & 8.33 & 6.59 \\
    Lane Change & 8.34 & 6.45 \\
    Stopped & 8.33 & 6.39 \\
    Stationary & 8.34 & 6.00 \\
    \hline
  \end{tabular}
  \caption{\label{tab:maneuver_distribution}
    Maneuver type distribution of $D_{base}$ showing original uniform tags vs processed tags with visibility logic applied.
  }
\end{table}

\subsection{Manual Review}

A team of human expert labelers manually reviewed 360 randomly sampled MCQs from the test set of $D_{llm}$ after Stage~II of data generation. Their answers are compared against the ground truth to obtain an accuracy score. These accuracy scores act as a baseline human evaluation score used for comparing model performance. The average accuracy among the expert labelers was 88.7\% with an inter-labeler agreement of 93.9\%. This demonstrates that the questions may be consistently answered between reviewers and provides an approximate baseline of accuracy numbers we might expect from fully-trained models. Models that outperform the human expert baseline may be doing so for a variety of reasons, such as a small sample size used to collect the baseline, input cueing, or learning patterns on incorrect labels.

\subsection{Results on Pretrained VLMs}
In addition to calculating accuracy of the pretrained models on the two datasets in Table \ref{tab:sanitycheck}, we also run shuffle evaluation without video. When we shuffle evaluate without video on pretrained models, we ideally expect close to 0 accuracy as the chance of picking the wrong answer in one of the four variants is high. We observe that $D_{llm}$ has higher performance suggesting more bias compared to $D_{new}$. Similar to the accuracy numbers in Table \ref{tab:sanitycheck}, Gemini 2.5 Pro which was used to create the data exploits the biases most. Additionally these biases again do not expose the complete extent of bias in the text as seen with shuffle results in Table \ref{tab:results}.

\begin{table}[h]
  \centering
  \begin{tabular}{llp{1.5cm}p{1.5cm}p{1.5cm}}
    \hline
    \textbf{Dataset} & \textbf{Model} & \textbf{Shuffle} \\
    \hline
    $D_{llm}$ & Gemini 2.5 Pro & 19.71 \\
    $D_{llm}$ & Gemini 2.5 flash & 9.79\\
    $D_{llm}$ & Qwen2-VL-7B & 13.86 \\
    \hline
    $D_{new}$ & Gemini 2.5 Pro & 8.68 \\
    $D_{new}$ & Gemini 2.5 flash & 5.57 \\
    $D_{new}$ & Qwen2-VL-7B & 6.40 \\
    \hline
  \end{tabular}
  \caption{\label{tab:sanitycheck_shuffle}
    Results obtained by passing the MCQ through LLMs without video context. 
  }
\end{table}

\subsection{Hyperparameters for training and inference}
The hyperparameters we used for our training runs are shown in Table \ref{tab:hyperparams}. We maintain consistent settings across all training runs, employing early stopping once the validation loss begins to increase. Extended training runs confirm the absence of the double descent phenomenon, justifying our use of early stopping at the first inflection point. Curriculum-specific configurations are applied only for curriculum learning runs. For evaluations we use greedy sampling, i.e. $k=1$ whenever possible. We train a Qwen2-VL 2B parameter model for ~20hrs on 8 B200 GPUs per training run.

\begin{table}[ht]
\centering
\begin{tabular}{ll} 
\textbf{Hyperparameter} & \textbf{Value} \\
\hline
\textit{Model Configuration} & \\
Base Model & Qwen2-VL-2B \\
Precision & bfloat16 \\
\hline
\textit{Optimization} & \\
Optimizer & AdamW \\
Learning Rate & $2 \times 10^{-5}$ \\
Weight Decay & 0.1 \\
Decay Type & Linear \\
Warmup steps & 100 \\
\hline
\textit{Training Setup} & \\
Global Batch Size & 256 \\
Random Seed & 42 \\
Max steps & 2500 \\
\hline
\textit{Curriculum Learning} & \\
$d_{min}$ & 0 \\
$d_{max}$ & 100 \\
$\tau$ & 670 \\
\hline
\end{tabular}
\caption{Hyperparameters for Qwen2-VL-2B Maneuver MCQ Fine-tuning. The curriculum learning parameter $\tau$ refers to the number of training steps over which curriculum is applied, $d_{min}$ and $d_{max}$ refer to the minimum and maximum drop rations.}
\label{tab:hyperparams}
\end{table}

\end{document}